\def\thesis{0}

\ifnum\thesis=0
\documentclass[10pt,twocolumn,letterpaper]{IEEEtran}
\else
\documentclass{article}
\usepackage[top=1in, bottom=1in, left=1in, right=1in]{geometry} 
\fi

\usepackage{times}
\usepackage{epsfig}
\usepackage{graphicx}
\usepackage{amsmath}
\usepackage{amssymb}

\usepackage[norule,symbol,perpage]{footmisc}
\usepackage[normalem]{ulem}
\usepackage{setspace}
\usepackage{times}
\usepackage{epsfig}
\usepackage{amsmath}
\usepackage{amssymb}
\usepackage{makecell}
\usepackage{verbatim}
\usepackage{balance}
\usepackage{color}
\usepackage{xspace}
\usepackage{enumerate}
\usepackage{graphicx}
\usepackage[noend]{algpseudocode}
\usepackage{algorithm}
\usepackage{wrapfig}
\usepackage{microtype}
\usepackage{pifont}
\usepackage{caption}
\usepackage{multirow}
\usepackage{cite}
\usepackage[bookmarks=false]{hyperref}
\usepackage{subcaption}

\usepackage{balance}
\usepackage[utf8]{inputenc} 
\usepackage[T1]{fontenc} 
\usepackage{csquotes}

\newcommand{\resist}{\textsc{Resist}}

\newcommand{\iriscode}{{Gabor filter template}}

\newcommand{\ignore}[1]{}
\usepackage{fancyhdr}

\begin{document}

	\title{Resist : Reconstruction of irises from templates}

\ifnum\thesis=0
\author{Sohaib Ahmad, \and Christopher Geiger, \and Benjamin Fuller \\
University of Connecticut\\
Storrs, CT USA\\
{\tt\small \{sohaib.ahmad,christopher.geiger,benjamin.fuller\}@uconn.edu}
}
\else
\author{Christopher Geiger\\
University of Connecticut\\
Storrs, CT USA\\
{\tt\small christopher.geiger@uconn.edu}
}
\fi

\graphicspath{ {images/} }

 \def\shownotes{1}
 \ifnum\shownotes=1
 \newcommand{\authnote}[2]{{\textcolor{red}{\textsf{#1 notes: }\textcolor{blue}{#2}}\marginpar{\textcolor{red}{\textbf{!!!!!}}}}}
 \else
 \newcommand{\authnote}[2]{}
 \fi
 \newcommand{\bnote}[1]{{\authnote{Ben}{#1}}}
 \newcommand{\snote}[1]{{\authnote{Sohaib}{#1}}}
 \newcommand{\cnote}[1]{{\authnote{Chris}{#1}}}

\newcommand{\cmark}{\normalsize \ding{51}}
\newcommand{\xmark}{\ding{55}}

\maketitle
\thispagestyle{empty}

\renewcommand{\refname}{References}

\begin{abstract}
Iris recognition systems transform an iris image into a feature vector.  The seminal pipeline \emph{segments} an image into iris and non-iris pixels, \emph{normalizes} this region into a fixed-dimension rectangle, and \emph{extracts} features which are stored and called a \emph{template} (Daugman, 2009). This template is stored on a system. A future reading of an iris can be transformed and compared against template vectors to determine or verify the identity of an individual.

As templates are often stored together, they are a valuable target to an attacker.  
We show how to invert templates across a variety of iris recognition systems. That is, we show how to transform templates into realistic looking iris images that are also deemed as the same iris by the corresponding recognition system. Our inversion is based on a convolutional neural network architecture we call RESIST (REconStructing IriSes from Templates).  

We apply RESIST to a traditional Gabor filter pipeline, to a DenseNet (Huang et al., CVPR 2017) feature extractor, and to a DenseNet architecture that works without normalization. Both DenseNet feature extractors are based on the recent ThirdEye recognition system (Ahmad and Fuller, BTAS 2019). When training and testing using the ND-0405 dataset, reconstructed images demonstrate a rank-1 accuracy of $100\%, 76\%$, and $96\%$ respectively for the three pipelines.  The core of our approach is similar to an autoencoder. However, standalone training the core produced low accuracy.  The final architecture integrates into an generative adversarial network (Goodfellow et al., NeurIPS, 2014) producing higher accuracy.
\end{abstract}


\section{Introduction}
\label{sec:intro}
This work\footnote{A preliminary version of this work was published at IEEE International Joint Conference on Biometrics in 2020~\cite{ahmad2020resist}.  In addition to substantive editorial changes, this manuscript additionally considers attack accuracy across datasets (Section~\ref{ssec:cross data set analysis}) and possible defenses (Section~\ref{sec:defense}).}  explores the vulnerability of storing the output of an iris recognition system. Irises are strong biometrics with a high entropy rate across individuals~\cite{itkis2015iris} and strong consistency of an individual's value over time~\cite{kronfeld1962gross,trokielewicz2020post}. The seminal processing pipeline due to Daugman proceeds in three stages~\cite{daugman2004iris}.  First, the iris region is \emph{segmented} from the rest of the image.  Second, this region is \emph{normalized} into a rectangular representation. Lastly, a \emph{feature extractor}, such as a 2-dimensional Gabor filter is applied.  The resulting values are converted to binary forming a template.   Daugman's pipeline is known as the iriscode.

Due to the noise experienced when collecting irises in real environments, new pipelines are steadily proposed.
In most systems, there are two main stored values. The first is the model or feature extraction mechanism (which may be a traditional set of feature extractors or a deep neural network architecture trained on irises).  The second is a set of output feature vectors known as \emph{templates} that are used to identify individuals.  When a person's biometric is collected, known as a \emph{reading}, the identity of this person is calculated (or verified) as the identity of the minimum distance stored iris.  In many applications for the identity to be considered a match, the distance must additionally be less than some predefined threshold. 

If an attacker has a target's iris they may be able to generate spoofed iris images~\cite{venugopalan2011generate,huang2018perspective,soleymani2019adversarial} to fool an iris recognition system. There are multiple ways a spoofed iris can be \emph{presented} to the iris recognition system, including a printed iris or a textured contact lens; these attacks are known as presentation attacks. 
Defenses to these attacks~\cite{connell2013fake,raja2016color,doyle2015robust} are usually trained on certain types of presentation attacks. As an example, Chen et al.~\cite{chen2018multi} use deep convolutional neural networks (CNNs) to detect presentation attacks. Kohli et al.~\cite{kohli2016detecting} extract Zernike moments and local binary pattern features which are used in a neural network classifier to detect presentation attacks.

A second attack vector inverts a template back into a corresponding an iris image~\cite{galbally2012iriscode}.  
An attack is successful if the resulting image is classified in the same class as the template by the recognition system. Templates are more accessible to an attacker than actual biometrics since recognition systems store templates to compare against any input to the system. Thus, if one can reconstruct an iris, one can execute a presentation attack in more settings.
Galbally et al.~\cite{galbally2012iriscode} presented such an inversion attack against an iris Gabor filter feature extractor.

While templates are usually encrypted at rest, for authentication systems in use, templates will be unencrypted in memory.
Template protection and cancellable biometrics can protect against inversion attacks threat~\cite{ratha2001enhancing,zuo2008cancelable}. Bloom filters~\cite{gomez2016unlinkable,bringer2015security,stokkenes2016multi} and fuzzy extractors~\cite{dodis2008fuzzy,juels1999fuzzy,juels2006fuzzy,boyen2004reusable,fuller2013computational} represent techniques to secure template storage~\cite{hernandez2009biometric,bringer2007optimal}. Recent research casts doubt into whether current techniques~\cite{canetti2021reusable} provide sufficient protection~\cite{simhadri2017reusable,keller2020fuzzy}.

\begin{figure*}[t]
    \begin{centering}
    \includegraphics[scale = 0.50]{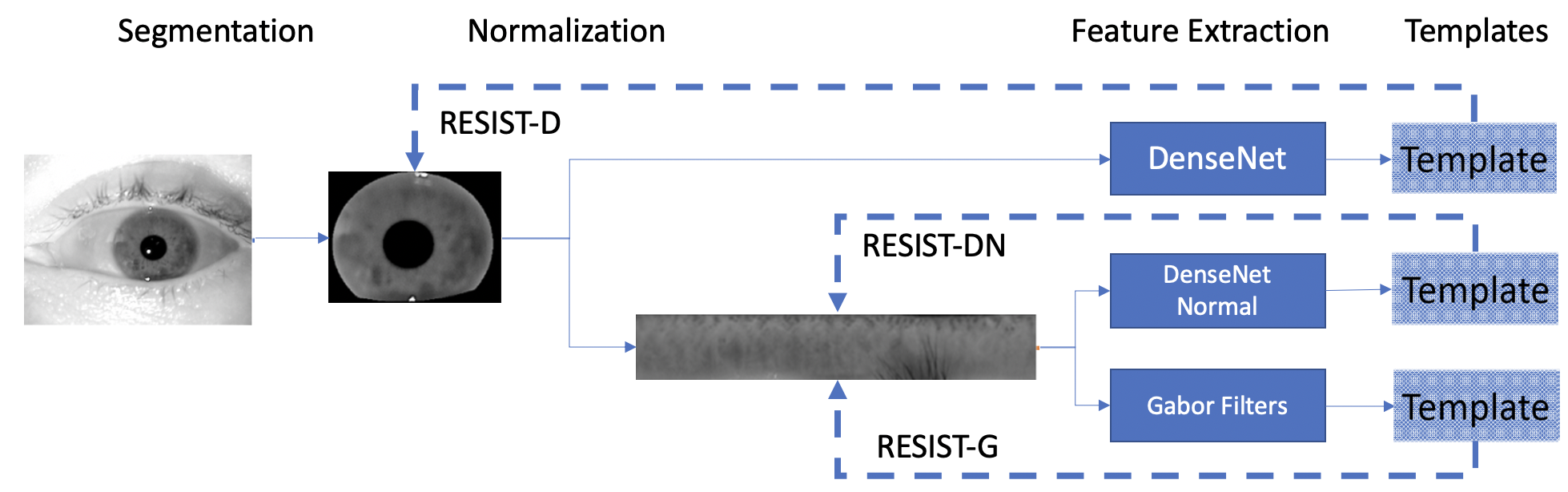}
    \caption{Our iris reconstruction system.  Blue lines represent 3 different iris processing pipelines.  The traditional Gabor filter pipeline is on the bottom. A DenseNet based pipeline on normalized images in the middle and a DenseNet pipeline on unnormalized images on the top.  Both DenseNet pipelines follow the three stages show in the top pipeline.  These are drawn from prior work and discussed in Section~\ref{sec:pipelines}.  For the two normalized pipelines, \resist{} reconstructs normalized images while for the pipeline without normalization it reverses to a segmented iris.}
    \label{fig:reconstruct_system}
    \end{centering}
\end{figure*}

\paragraph{Our Contribution}
The primary contribution of this work is a convolutional neural network architecture called \resist{} for REconStructing IriSes from Templates.  \resist{} effectively reconstructs irises from a stored template.  \resist{} is a black-box attack which does not utilize the specifics of the feature extraction to train the reconstruction network. However, we do assume knowledge of the length of the feature vector. We apply \resist{} to three iris processing pipelines which are summarized in Figure~\ref{fig:reconstruct_system}:
\begin{description}[\IEEEsetlabelwidth{\resist{}-DN}]
    \item[\resist{}-G] A traditional Gabor filter based processing pipeline (with segmentation and normalization before applying Gabor filters).
    \item[\resist{}-DN] A deep neural network feature extractor based on DenseNet~\cite{huang2017densely} that works on normalized images.
    \item[\resist{}-D] A deep neural network feature extractor based on DenseNet that directly works on segmented images. 
\end{description}

\noindent
Both deep neural network approaches build on the recent ThirdEye architecture~\cite{ahmad2019thirdeye}, see Section~\ref{sec:pipelines}.  
Our goal is to directly invert the resulting feature vector. For the neural networks this is a vector of real values. It is a binary vector for \resist{}-G.  In many applications, the real valued vector resulting from a neural network is projected to a binary vector using locality sensitive hashes~\cite{jin2017ranking}, we leave inverting this system as future work. Non-iris pixels are removed by the segmentation layer and should not affect the resulting template, so \resist{} only attempts to find the segmented image.  For the two pipelines (\resist{}-DN and \resist{}-G) that explicitly normalize we invert to a normalized image for two reasons: 1) normalization may also lose information, such lost information clearly can not be recovered from a template, and 2) to focus on the privacy implications of the feature extractor.

\paragraph{Technical Approach}
The \emph{core} of \resist{}'s technical network is similar to an autoencoder.  However, training an autoencoder core on its own yielded low accuracy.  To improve accuracy, we use techniques from the area of synthetic iris generation.  Recent machine learning techniques such as generative adversarial networks or GANs~\cite{goodfellow2014generative} have made it possible to generate synthetic irises given access to a database.  A GAN trains two networks in competition, a generator which should produce synthetic irises and a discriminator which classifies irises as real or synthetic.  Yadav et al.~\cite{yadav2019synthesizing} uses RaSGAN (relativistic average standard GAN)~\cite{jolicoeur2018relativistic} to generate synthetic irises for the purpose of studying their effects on presentation attack detection (PAD) algorithms. Irises from the RaSGAN perform well against PAD and follow real iris statistics well. Kohli et al.~\cite{kohli2017synthetic} use the DCGAN architecture to generate synthetic irises.

Since our standalone \resist{} inversion core is not accurate, as a second stage we train our core as the \emph{generator} of the GAN.  The discriminator then attempts to distinguish between real and inverted irises.  The synthetic iris database task learns a generator $g$ that takes randomness $r$ to produce an iris that cannot be distinguished from a legitimate image.  Synthetic irises can be viewed as irises that must closely resemble bonafide irises as discussed in~\cite{yadav2019synthesizing,kohli2017synthetic}.
Iris reconstruction adds an input to the generator: which template the synthetic iris should match.  Thus, reconstructing irises from corresponding templates is a harder task.  It is not enough to learn the distribution of the iris.  \resist{} must learn from the individual template. 

\begin{figure}[t]
    \begin{centering}
    \includegraphics[scale = 0.55]{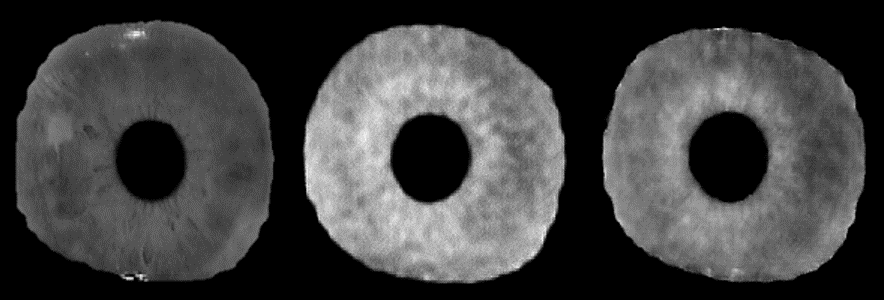}
    \caption{Iris reconstructions from templates. Original, network core and GAN reconstructions respectively. }
    \label{fig:reconstruct_type}
    \end{centering}
\end{figure}

Figure~\ref{fig:reconstruct_type} shows a standard iris image on the left.  As mentioned above, our approach has two training stages, first we build a core that is similar to an autoencoder. This core is then placed inside of a GAN for a second training stage.  The center image is a reconstructed iris using the core, this image has low pixel error (optimized by the core) with an average pixel error of $3\%$.  However, when matched with real iris images, this technique only achieves rank-1 accuracy of $62\%$ and true acceptance rate (TAR) of only $18.5\%$ at 1\% false acceptance rate (FAR). Rank-1 accuracy is how frequently an iris template of the same class is the closest value in the dataset.  

As mentioned above, to deal with this inaccuracy, we introduce the second stage of training, the core is moved inside a GAN as the generator.  When trained from scratch, this GAN collapsed, that is, it always produced the same image.   To prevent collapse, we use three techniques:
\begin{enumerate}
\itemsep0em
\item Pretraining the core before training in the GAN (pretraining was used in prior iris recognition networks~\cite{boyd2020deep}), 
\item Adding noise to the real iris images during training, and 
\item Using spectral normalization~\cite{miyato2018spectral}. 
\end{enumerate} 
The right hand image in Figure~\ref{fig:reconstruct_type} shows a characteristic image output by \resist{}.

\paragraph{Overview of Results}
 We report rank-1 accuracy as well as accuracy metrics from a recent facial reconstruction paper~\cite{mai2018reconstruction}. We report this for both the  \resist{} templates and the corresponding legitimate templates. The ND-0405 dataset is used for training and testing.  We summarize our findings here with the rank-1 accuracy of reconstructed irises compared with rank-1 accuracy of legitimate irises (see Table~\ref{tab:resist}):
\begin{description}[\IEEEsetlabelwidth{\resist{}-DN}]
\item[\resist{}-D] $96.3\%$ vs. DenseNet: $99.9\%$,
\item[\resist{}-DN] $89.3\%$ vs. DenseNet Normal: $98.7\%$, and
\item[\resist{}-G] $97.1\%$ vs. Gabor filter: $97.9\%$.
\end{description}

\noindent
The underlying system accuracy is an upper bound for our accuracy (as the reconstructed template is passed through the underlying feature extractor). In a preliminary investigation, we applied the trained \resist{} models to feature extractors trained for the CASIAv4 Interval~\cite{casia} and IITD datasets~\cite{kumar2010comparison}.  While $\resist{}$ recovered some structural information about the shape of the iris, it did not recover fine grained textural details.
We explore the accuracy rates further in Section~\ref{sec:results}.  

Model inversion is a closely related task that reconstructs inputs from classification vectors.  Prior work in model inversion has shown regularization to be an effective defense~\cite{shokri2017membership,yang2020defending}.  While the trained feature extractors already used regularization, adding additional regularization does mitigate the effectiveness of $\resist{}$. This defense does come at a small cost to accuracy.  See Section~\ref{sec:defense} for more detail.  

\paragraph{Related Work}
Galbally et al.~\cite{galbally2012iriscode} use a genetic algorithm which reconstructs irises from their Gabor filter templates. Their method is stochastic and generates multiple irises from a single template. Venugopalan et al.~\cite{venugopalan2011generate} show how given an Gabor filter template $f(x)$ for an individual $x$, another individual $y$ can transform their iris image into one with a similar \iriscode as $x$.
Mai et al.~\cite{mai2018reconstruction} recently considered reconstruction attacks on the facial biometric. Their attack relies on a large training dataset containing 2 million facial images. To our knowledge such a large iris dataset does not exist. Facial recognition is done using deep neural networks which work on a centered face image to generate a template. There are multiple ways to generate templates from the iris biometric. We primarily explore the different types of iris templates and use a single reconstruction network in contrast to Mai et al. where a GAN is used to augment their existing training dataset and train a reconstruction network on the augmented facial images. To the best of our knowledge, \resist{} is the first attack that reconstructs irises from deep learning based processing templates.  

\paragraph{Organization}
The rest of this work is organized as follows.  In Section~\ref{sec:pipelines} we describe the three attacked pipelines, Section~\ref{sec:design} describes the design of \resist{}, Section~\ref{sec:evaluation} provides an overview of the evaluation strategy and the dataset used with results in Section~\ref{sec:results}, Section~\ref{sec:defense} considers countermeasures.  Section~\ref{sec:conclusion} concludes.

\section{Iris Processing Pipelines}
\label{sec:pipelines}
We apply RESIST to three iris processing pipelines.  This section briefly introduces the attacked pipelines.
We explore both Gabor filter and deep learning based systems. 
In traditional iris recognition systems there are three stages:
\begin{description}[\IEEEsetlabelwidth{Feature Extraction}]
\item[Segmentation] classifies each pixel as iris or non iris,
\item[Normalization] converts the iris region into a fixed-area rectangle,
\item[Feature Extraction] distills the rectangle into features.
\end{description}

\noindent In the traditional \iriscode{} pipeline feature extraction consists of two stages 1) convolving small regions with Gabor filters and 2) converting the resulting complex numbers to bits by taking the sign. Subsequent iris readings by the system undergo the same procedure.  The similarity of these binary templates is computed using the Hamming distance. Our goal is to take the binarized template and reconstruct the normalized iris representation. We form a traditional iris recognition pipeline using USIT software to do segmentation and normalization~\cite{hofbauer2014ground}. We use transform 5 from Osiris's~\cite{othman2016osiris} Gabor filter banks for feature extraction. 

We consider two deep-learning based iris pipelines based on the ThirdEye~\cite{ahmad2019thirdeye} iris processing system. We use the NotreDame 0405~\cite{bowyer2016nd} dataset to train these two models. Both DenseNet networks use the first 25 left iris images of each subject for training. The test comprises first 10 right iris images of each subject. Both DenseNet networks are trained to convergence (98\% rank-1 accuracy on testing set). 
We further report on the accuracy of these models in Section~\ref{sec:results}. 
For both pipelines, the training process of the DenseNet is to the specification of ThirdEye~\cite{ahmad2019thirdeye}.
 However we replace the ResNet~\cite{he2016deep} with DenseNet using DenseNet-169 as our feature extractor. The ResNet based architecture achieves an EER (equal error rate) of 1.32\% on the ND-0405 test set specified by Ahmad and Fuller whereas changing to a DenseNet based architecture improves the EER to 1.16\%. The resulting EER is close to state of art EER of 0.99\%~\cite{zhao2017towards}. The DenseNet learns feature embeddings per iris by minimizing the triplet loss~\cite{schroff2015facenet}. The trained network outputs a 1024 dimensional feature vector per iris. A simple augmentation of flipping the iris image along the horizontal axis is done and another feature vector is extracted. This augmentation improves the recognition accuracy yielding a feature vector of 2048 dimensional feature vector per iris. 

The first pipeline uses normalized iris images obtained by using the USIT software~\cite{hofbauer2014ground}. This feature vector serves as the input for \resist{}-DN.  
 
The second pipeline does not include an explicit normalization step.  Instead it uses a two stage pipeline using the segmentation pipeline of Ahmad and Fuller~\cite{ahmad2018unconstrained} as direct input to the ThirdEye network. Ahmad and Fuller's segmentation accuracy is comparable to state of art~\cite{wang2020towards}. We call these two pipelines DenseNetNormal and DenseNet respectively. These three pipelines demonstrate a variety of representative techniques (Gabor filters, normalization and CNNs) used in iris recognition.


\section{Design}
\label{sec:design}
The \emph{core} of \resist{} is a convolutional network similar to an autoencoder. We train the core standalone and inside a GAN as the generator.  
The objective of an autoencoder is to compress its input to form a feature representation and then recreate the input from the compressed feature representation. That is, the autoencoder learns two functions $f$ and $f'$ where the range of $f$ has smaller dimension and $f'(f(x)) \approx x$ for all trained values.

We slightly abuse this framework; we use templates as the input value $x$ and try to reconstruct iris images $\hat{y}$. Our core is based on U-Net~\cite{ronneberger2015u} which has been used in pix2pix GANs~\cite{isola2017image} for image to image translation and fits our template inversion use case. The full architecture is presented in Section~\ref{ssec:arch}. 
We first describe the loss functions for the network core when trained standalone.  We then describe integration in the GAN, and finally present the overall architecture.

\subsection{Standalone Network Core Training}
\label{ssec:standalone}

Our \emph{standalone} core training works in two stages. First, we train the core using the L1 loss. Our models minimize the absolute differences (of pixels) between the predicted and the ground truth iris image. The L1 loss or mae~(mean absolute error) provides crisper details in reconstructed images than L2 loss~\cite{wang2018esrgan}. The L1 loss function: 
\begin{equation}
L_{mae} = \frac{\sum_{h,w} \lvert y(h,w) - \hat{y}(h,w)\rvert}{h*w},
\label{l1}
\end{equation}

\noindent
where $h,w$ denote the height and width of the image, $y$ is the original image and $\hat{y}$ is the output of the core. 
To fine tune the reconstructions we add two new losses. The first optimizes the Structural similarity (SSIM)~\cite{wang2004image} between images.  Structural similarity measures the product of two terms.  The first term of SSIM is roughly the product of the two image means normalized by the sum of means.  The second term is the covariance of the two images normalized by the sum of variance.  The actual definition adds an additional term in the normalization to account for small values.  We minimize structural dissimilarity between the real and reconstructed images:

\begin{equation}
L_{SSIM} =  1-SSIM(y,\hat{y})
\label{SSIM}
\end{equation}

The third loss function is the Perceptual loss~\cite{johnson2016perceptual}. This loss takes as input the reconstructed image from the first head (T2 in Figure~\ref{fig:gan_arch}) and passes it through a pre-trained VGG16~\cite{Simonyan15} network. This VGG16 network is trained on the ImageNet dataset~\cite{deng2009imagenet} and not fine tuned on any iris dataset. The VGG16 network serves as a texture based feature extractor where an intermediary layer represents textures in the reconstructed iris image. Perceptual loss minimizes the L2 distance between textures of the real and reconstructed iris images. 
\vspace{-.1in}

\begin{equation}
L_{Perceptual} = \frac{ (\lvert \phi_j(y) - \phi( \hat{y} )\rvert)}  {C_j H_j W_j}
\label{perceptual}
\end{equation}

\noindent
where $\phi$ denotes activations of VGG16 after its 9th convolution operation, C,H,W are layer activation dimensions and $y$ and $\hat{y}$ are real and reconstructed iris images.

\begin{figure}[t]
    \begin{centering}
    \includegraphics[scale = 0.45]{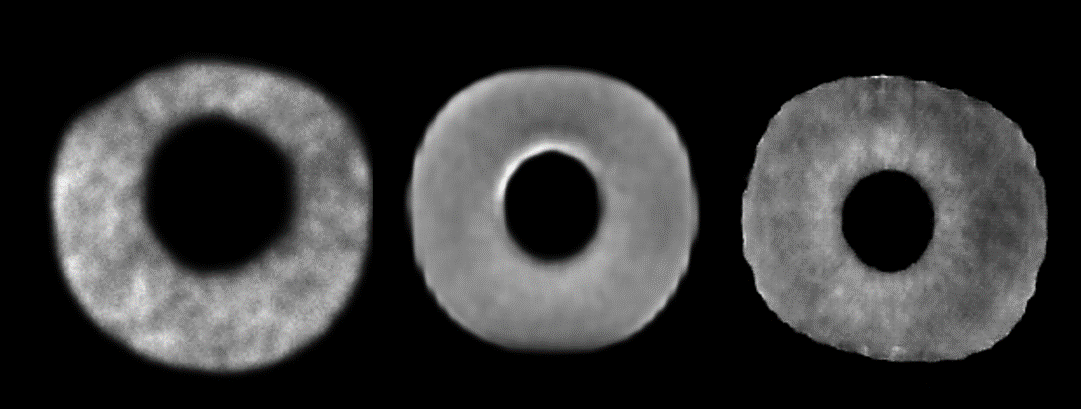}
    \caption{Iris reconstructions after addition of L1 loss, SSIM loss and perceptual loss from left to right. }
    \label{fig:reconstruct_heads}
    \end{centering}
\end{figure}
\begin{figure}[t]
    \begin{center}
    \includegraphics[scale = 0.43]{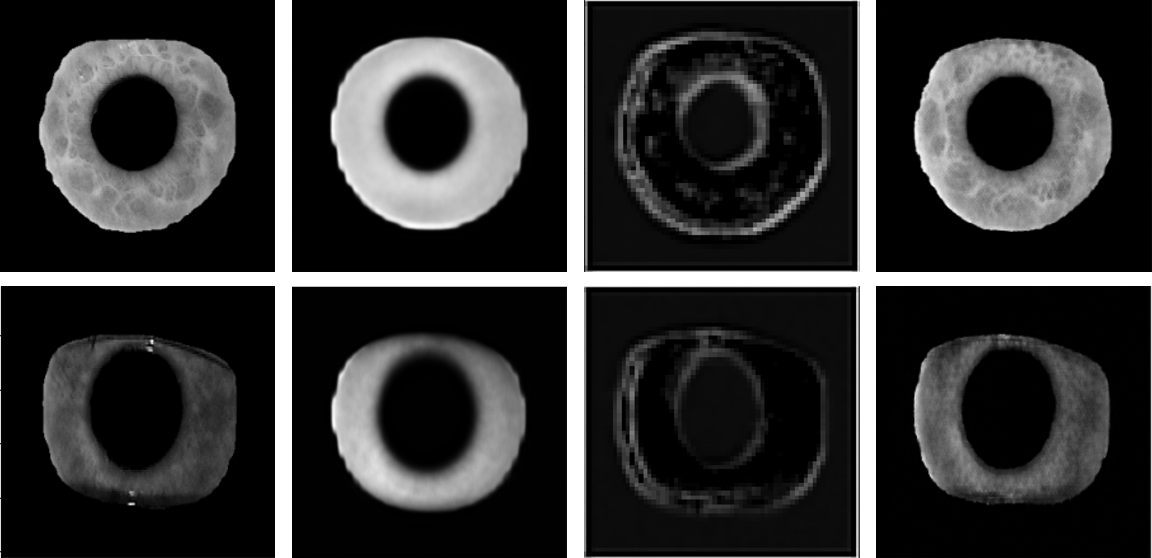}
    \caption{The contribution of different loss functions visually shown. The first column  is an original image.  The second and third columns are the image contributions provided by the SSIM and perceptual loss functions respectively. The last column shows the reconstructed images. }
    \label{fig:loss_images}
    \end{center}
\end{figure}

\noindent
\textbf{The need for the three loss functions}
To illustrate how the three loss functions work to improve the core, we discuss accuracy numbers of this standalone core using the DenseNet pipeline as an example. Reconstructions using just $L_{mae}$  capture the boundary and some high level textures of the iris.  However, such images have an average L1 distance of 8\%. Furthermore, the quality of these images is not high, the left of Figure~\ref{fig:reconstruct_heads} shows a reconstruction using just $L_{mae}$. Adding $L_{SSIM}$ improves the image quality and actually reduces $L_{mae}$ to $4\%$. $L_{SSIM}$ improves the reconstructions of the first head by extracting features from the output of the first head and trying to reconstruct an iris image by minimizing structural dissimilarity. The reconstructed irises now contain better textured iris patterns however they are not sharp relative to their equivalent real irises. This can be seen in the center iris of Figure~\ref{fig:reconstruct_heads}.  Finally, we employ $L_{Perceptual}$ loss~\cite{johnson2016perceptual}. Average L1 distance after using perceptual loss drops to $3\%$. Furthermore, the texture quality improves as well, see the right image in Figure~\ref{fig:reconstruct_heads}.  This is also shown explicitly in Figure~\ref{fig:loss_images} which shows the image contributions of the various loss functions.  The SSIM loss function provides the general shape and structure while the perceptual loss is critical to providing fine grained detail.

To summarize, the standalone training for the core has two phases, first the core is trained on just the L1 loss to capture the iris structure. The second phase adds two losses and one extra head to the core. The two heads are T2 and T3 in Figure~\ref{fig:gan_arch}. Both $L_{mae}$ and $L_{Perceptual}$ are on the T2 head. The T3 head minimizes $L_{SSIM}$. The standalone core is trained to minimize the loss:
\begin{equation}
L_{AE}= \alpha L_{Perceptual} + \beta L_{mae} + \gamma L_{SSIM}
\end{equation}
Where $\alpha,\beta,\gamma$ are weight terms for the equation with values 2,1,1 respectively.

\subsection{Improved Core Training in a GAN}
We discussed training the core in a standalone fashion. Now we train the core as a generator inside a GAN. The idea for this comes from prior work which shows training autoencoders in an adversarial setting yields better results than training the standalone~\cite{makhzani2015adversarial}. 

\begin{figure*}[t]
    \begin{center}
    \includegraphics[scale = 0.50]{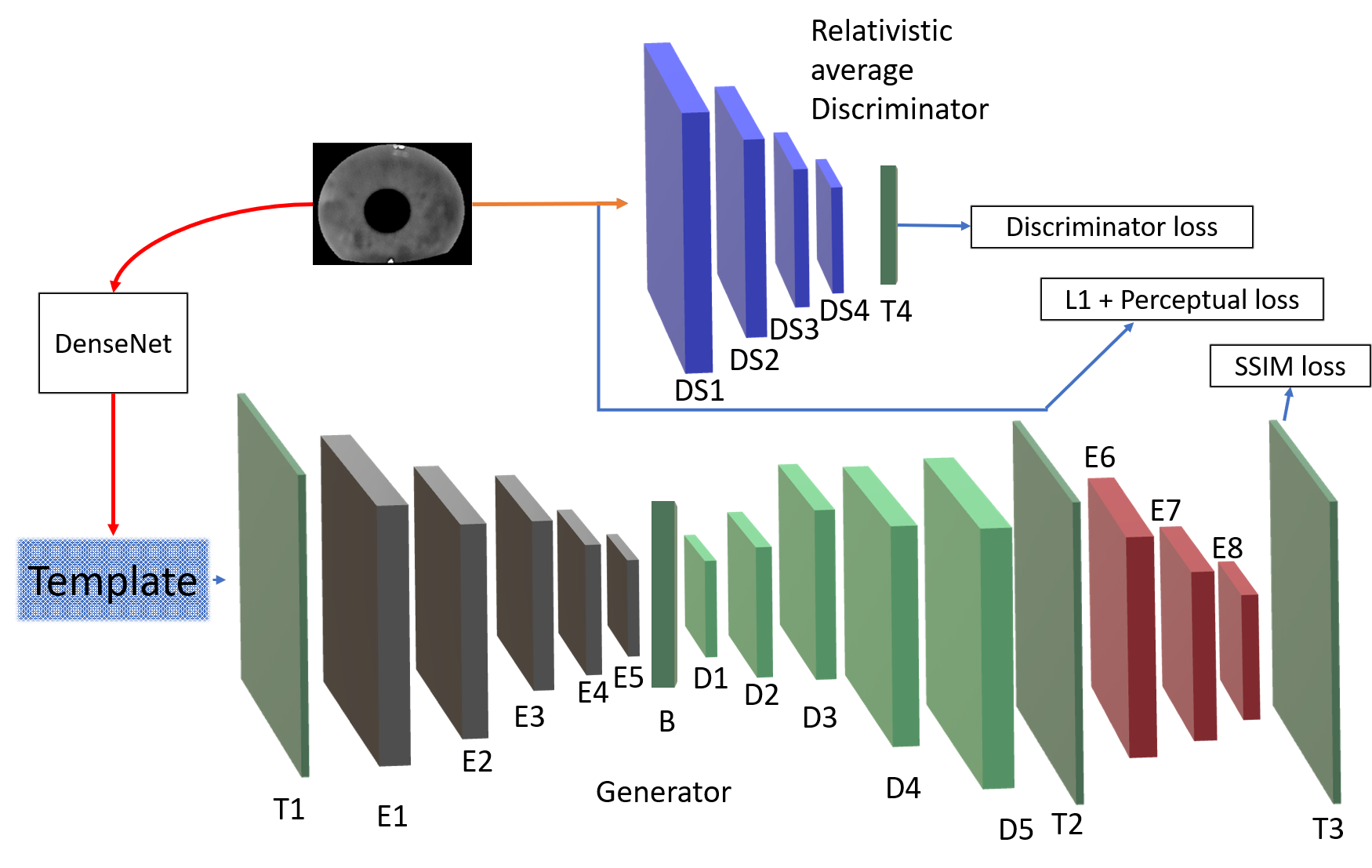}
    \caption{\resist{} architecture. The generator reconstructs iris images from its corresponding template. The discriminator judges the reconstructed iris as real or fake. The DenseNet is not part of \resist{} training. Perceptual loss is minimized by passing reconstructed iris from T2 to a VGG16 network which is not shown in the figure. }
    \label{fig:gan_arch}
    \end{center}
\end{figure*}

The final architecture is a GAN which takes as input a template and yields a reconstructed iris image (see Figure~\ref{fig:gan_arch}). In essence when \resist{} is coupled with a feature extractor the composition of the two models acts as an autoencoder.  The iris image is converted to a template by the feature extraction network and back to an iris image by \resist{}. A GAN has a generator which generates images and a discriminator which judges how good the generated images are. Our generator is the core discussed above. We use a recently proposed relativistic average discriminator~\cite{jolicoeur2018relativistic} as our discriminator. To build up to the relativistic discriminator we first start with the original GAN loss functions: 
\vspace{-.1in}

\begin{align}
L(D) =& -\mathbb{E}_{y \sim \mathbb{P}_{real}} [\log(D(y))]\\ &-\mathbb{E}_{\hat{y} \sim \notag \mathbb{P}_{{fake}}} [\log(1 - D(\hat{y}))] \\
L(G) =& -\mathbb{E}_{\hat{y} \sim \mathbb{P}_{fake}} [\log(D(\hat{y}))].
\end{align}


\noindent
 $L(D)$ is called the discriminator loss and $L(G)$ is called the generator loss. $y$ and $\hat{y}$ are the original and reconstructed irises. 
 
 In the original GAN formulation a noise vector is sampled from a multivariate normal distribution with a mean of 0 and a variance of 1. This vector is transformed into an image by the generator while the discriminator compares the generated image and the ground truth image. $\mathbb{P}_{real}$ is the distribution of real images and $\mathbb{P}_{fake}$ is usually a multivariate normal distribution with mean of 0 and variance of 1, $\mathbb{P}_{fake}$ is the distribution of the iris templates for \resist{}.
 
 Both losses are minimized using gradient descent. The generator and discriminator play a zero sum game.  The generator weights are updated based on how good its fake images are while the discriminator weights are updated on how well it differentiates between real and fake images. A discriminator outputs a probability, therefore the last layer of the discriminator is a sigmoid activation. To achieve this we add a sigmoid function $\sigma$ as the last layer of a discriminator, we then define a discriminator $D$ as:

\begin{equation}
    D(y): \sigma(C(y))
\end{equation}
Where $\sigma$ is the sigmoid function and $C$ is a layer in any discriminator preceding the final sigmoid output. This layer is layer T4 in our architecture shown in Figure~\ref{fig:gan_arch}. $C(x)$ is thus the non-transformed discriminator output (before the sigmoid function). The discriminator will ideally output $1$ for real images and $0$ for fake images.

We use the relativistic average discriminator in our work which judges the generated images relative to their real counterparts and vice versa. This means that:

\vspace{-3mm}
\begin{align}
D_{Ra}(y)= \sigma(C(y-\mathbb{E}[C(\hat{y}]))\\
D_{Ra}(\hat{y})= \sigma(C(\hat{y}-\mathbb{E}[C(y]))\\
& \notag
\end{align}

\vspace{-3mm}
Where $\mathbb{E}$ is the expectation across all fake data samples in a batch.
With the definition of a discriminator above, the two original loss functions become:

\vspace{-3mm}
\begin{align}
    L(D_{Ra}) =& -\mathbb{E}_{y \sim \mathbb{P}_{real}} [\log(D_{Ra}(y))] \\
    &-\mathbb{E}_{\hat{y} \sim \notag \mathbb{P}_{fake}} [\log(1 - D(\hat{y}))] \\ 
    L(G_{Ra}) =& -\mathbb{E}_{\hat{y} \sim \mathbb{P}_{fake}} [\log(D_{Ra}(\hat{y}))] \\
    &-\mathbb{E}_{y \sim \mathbb{P}_{real}} [\log(1 - D(y))]\notag 
\end{align}

\vspace{-1mm}
\noindent
$\hat{y}$ is the generated output from the convolutional network core and $y$ is the real iris image. 
The generator takes templates as input to generate reconstructed irises, $\mathbb{P}_{fake}$ is therefore the distribution of the templates and $\mathbb{P}_{real}$ is the distribution of the real iris images. The final loss for the generator of all variants of \resist{} becomes:
\begin{equation}
L(G_{\resist{}}) = L(G_{Ra}) + \alpha L_{Perceptual} + \beta L_{mae} +\gamma L_{SSIM}
\end{equation}

To summarize, all \resist{} variants are trained in two stages. The core is trained standalone using the process described in Section~\ref{ssec:standalone}. This trained core is then used as the generator of a \resist{} network which is trained as a GAN. Both training stages use the Adam optimizer~\cite{kingma2015adam}.

\textbf{Training}
In addition to using the relativistic discriminator and using an already trained generator we employ different techniques to stabilize the training of \resist{}. Spectral normalization~\cite{miyato2018spectral} is applied to both discriminator and generator. Convolution and dense operators are both normalized using their spectral norms. Spectral normalization provided sufficient stabilization for \resist{} training, hence other stabilization techniques were not used. We also use two time scale update rule~\cite{Heusel2017GANsTB} for training, our learning rates are 1x$10^{-5}$ and 1.5x$10^{-5}$ for the generator and discriminator respectively. Gaussian noise sampled from a normal distribution of mean 0 and variance 1 is added to every pixel of real irises during training. \resist{} variants are trained for 400 epochs (200 steps per epoch) with a batch size of 12. Each batch consists of $y$ and $\hat y$ pairs.

For \resist{}-D and \resist{}-DN the template is the output of the DenseNet based feature extractor and the DenseNet Normal feature extractors described in Section~\ref{sec:pipelines}. From an architectural point of view the only difference is the output dimension.  \resist{}-G differs from both \resist{}-D and \resist{}-DN in input templates because it is binary (instead of real-valued).

\subsection{Architecture}
\label{ssec:arch}
\resist{} architecture borrows concepts from U-Net~\cite{ronneberger2015u} which has skip connections~\cite{he2016deep} from all its encoder layer to its decoder layers. The use of skip connections allows more gradient to flow through the network. 

The layer and architecture details are in Tables~\ref{tab:layer} and~\ref{tab:arch} respectively.  The architecture is shown using DenseNet as an example pipeline in Figure~\ref{fig:gan_arch}. First layer~(T1) transforms the template to the same dimension of the iris image. This is done using a fully connected layer. The encoder consists of five layers~(E1-5 in Table~\ref{tab:arch}), each layer has a convolution operation with a LeakyReLU~\cite{maas2013rectifier} activation followed by batch normalization~\cite{10.5555/3045118.3045167}. Dimensionality is reduced by strided convolution. The decoder also has five layers~(D1-5 in table~\ref{tab:arch}), each layer has deconvolution operation with a ReLU activation followed by batch normalization. Each decoder layer accepts a skip connection from its corresponding encoder layer, as an example encoder layer 5's gradient will be added to decoder layer's 1's gradient and so on for other layers in the encoder-decoder architecture in a U-Net fashion. Dimensionality is regained by strided deconvolution.

\begin{table}
\footnotesize
	\begin{centering}
		
		\begin{tabular}{|l|c|c|}
			
			\hline
			Layer&Elements\\
			\hline
			T1&Dense \\
			E*&Conv(Stride 2),BN,LeakyReLU \\
			D*&DeConv(Stride 2),BN,ReLU,Skip \\
            T2&Conv\\
            T3&Dense,Conv\\
            DS*&Conv(Stride 2)\\
            T4&Dense\\

			\hline
	\end{tabular} 
	
	    \end{centering}
\caption{Layers}
\label{tab:layer}
\vspace{-.1in}
\end{table}

\begin{table}
\footnotesize
\centering
		
		\begin{tabular}{|l|c|c|}
			
			\hline
			Layer&OutputSize&Kernels\\
			\hline
			Input&1x1024&-  \\
			T1&1x256x256&1  \\
			E\{1-5\}&Input Size/2&32,64,128,256,512,64,128,256 \\
			D\{1-5\}&Input Size*2&512,256,128,64,32 \\
			T2&1x256x256&1  \\
			T3&1x256x256&1  \\
            DS\{1-4\}&Input Size/2&128,128,128,128\\
            T4&1&1  \\

			\hline
	\end{tabular} 
\caption{GAN architecture}
\label{tab:arch}
\vspace{-.1in}
\end{table}

The first head of \resist{} reconstructs the iris image using a convolution operation coupled with a sigmoid activation~(T2) while minimizing the L1 loss from equation~\ref{l1}. This reconstructed image is passed through a VGG16~\cite{Simonyan15} network to have its perceptual loss minimized in equation~\ref{perceptual}. The reconstructed image from the first head is also passed through another series of encoder layers and another reconstructed iris image is formed. These encoder layers accept skip connections from the encoder-decoder. This second reconstruction is formed by using a dense layer with pixels equivalent to the dimension of the iris image followed by a convolution operation coupled with a sigmoid activation~(T3) which minimizes the loss in equation~\ref{SSIM}.


The reconstructed image from the first head is output and fed to the discriminator along with its corresponding real image. The discriminator judges the reconstruction relative to the real iris image. The discriminator has four layers each comprising strided convolution operations~(DS1-4). After the fourth convolution the features are flattened and passed to a dense layer which has a sigmoid activation outputting a probability~(T4).



\section{Evaluation}
\label{sec:evaluation}

\label{sec:evaluation}
All models are trained on the ND-0405~\cite{bowyer2016nd} dataset (we study \resist{} accuracy when trained on ND-0405 and tested on two other datasets in Section~\ref{ssec:cross data set analysis}). The ND-0405 dataset contains 64,980 iris samples from 356
subjects and is a superset of the NIST Iris Challenge Evaluation dataset~\cite{phillips2008iris}. Iris images are captured using the LG 2200 (Near infrared) biometric system. The images have blurring from motion and some out of focus images.

\paragraph{Methodology}

We described how the two DenseNet models are trained in Section~\ref{sec:pipelines}. All three \resist{} variants are trained on  features from all remaining left iris images from each subject~(apart from the 25 images per subject used to train the feature extractors). The testing set for all three pipelines is $20\%$ of right iris images. The training and testing set is kept same for all three pipelines. We consider the right and left irises of a subject as separate classes based on prior work showing statistical independent~\cite{daugman2004iris}. Training and testing sets are class disjoint. Furthermore, no training images for \resist{} are used in training the DenseNet models. 

\noindent
\textbf{DenseNet}
All images used to train DenseNet feature extractor and train \resist{}-D have been segmented using a publicly available Mask R-CNN based iris segmentation tool~\cite{ahmad2018unconstrained}. The feature extractor (DenseNet) is trained on these images while minimizing the triplet loss and generating feature embeddings per iris.
The segmented irises do not contain any non-iris occlusions. The segmented images are centered and resized to a resolution of 256x256 and directly fed to the feature extractor without any pre-processing. 

\noindent
\textbf{DenseNetNormal} For \resist{}-DN the images are segmented and normalized using the USIT software. The images are of resolution 64x512. These images contain some non-iris occlusions (eyelid, eyelashes). We do not use the masks provided per iris image by the USIT software. The normalized iris images are directly fed to the normalized iris feature extractor. 

\noindent
\textbf{Gabor Filter}
For the Gabor filter feature extractor, we use filter bank 5 from OSIRIS. The binarized output after convolution of normalized iris image with filter bank 5 of OSIRIS is our template. This is fed to \resist{}-G which learns to reconstruct back the corresponding normalized iris image. 



We explore two types of attacks defined by Mai et al.~\cite{mai2018reconstruction}, \textbf{Type-1} attack where a reconstructed iris is matched with its corresponding real iris.  \textbf{Type-2} attacks where reconstructed irises are matched with real irises of the same class excluding the image used to create the template. Type-1 and Type-2 accuracy depend on a distance threshold.  This distance threshold is varied to trade off between TAR/FAR. When we vary between TAR/FAR we are changing the threshold according to the feature extractors and observing the resulting change on \resist{}. We see the resulting ROC curves in Figure~\ref{fig:roc}.  We also report \textbf{Rank-1} of our reconstructed irises. Rank-1 accuracy is an all-all matching and does not use a threshold. A reconstructed iris has its distance calculated across the entirety of the real dataset. A true positive in this case is if the lowest distance (between a reconstructed and a real iris) is with an iris of the same class. We use the cosine distance for our deep templates and the Hamming distance for \iriscode{}s.


\section{Results}
\label{sec:results}
\begin{figure*}[h]
    \begin{centering}
    \includegraphics[scale = 0.53]{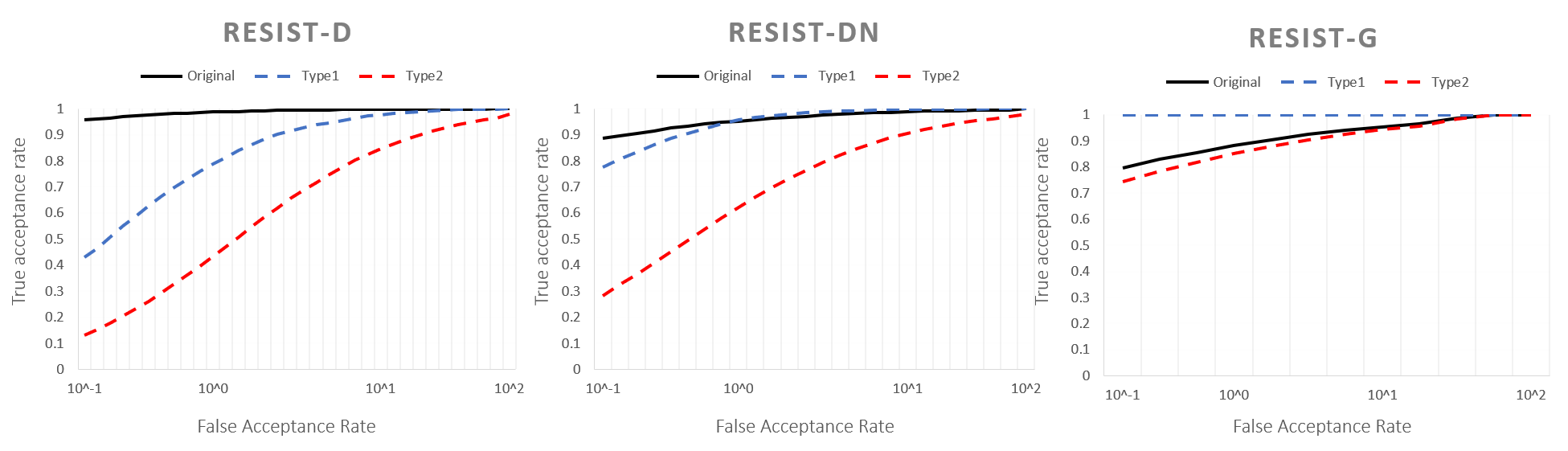}
    \caption{ROC curves for our three reconstruction pipelines. The x-axis is the false acceptance rate and the y-axis is the true acceptance rate. The Original curve is Type-2 error for the legitimate biometrics.  For legitimate systems there is no Type-1 error.}
    \label{fig:roc}
    \end{centering}
\end{figure*}

We summarize our results in Table~\ref{tab:resist} and Figure~\ref{fig:roc}.  In Table~\ref{tab:resist}, the legitimate row describes our three accuracy measures on legitimate irises.  Note that Type-1 accuracy will always be $100\%$ for a deterministic, legitimate transform (as the exact same template will be produced twice).

The first reconstruction is done using \resist{}-D.  The DenseNet based feature extractor is robust, achieving 99.9\% rank-1 accuracy on legitimate irises on the test set. \resist{}-D achieves a rank-1 accuracy of 96.29\% when matching the reconstructed test set against the real test set. This means that in a recognition system a reconstructed iris would be classified as a member of the stored database 96\% of the time a match is made. TAR at 1\% FAR is 43.3\% which represents the Type-2 accuracy rate for \resist{}-D. Type-1 accuracy for \resist{}-D is 76\%.  We also show results from our standalone network core.


Template inversion for deep templates for normalized irises works better than irises without normalization. The Rank-1 accuracy is lower at 89\% for \resist{}-DN vs 96\% for \resist{}-D. We attribute this to the normalized feature extractor model having worse accuracy compared to the segmented iris feature extractor (98\% vs 99\%). We also do not use any masks from the USIT software therefore the feature extractor has to learn extra information about occlusions. Eyelashes and eyelids are some common occlusions found in normalized iris images. \resist{}-DN has higher Type-1 and Type-2 accuracy rates than \resist{}-D. The Type-1 accuracy rate is especially high at 96\% when compared to the Type-1 accuracy of \resist{}-D at 76\%.

Previous works have shown that \iriscode{} based templates can be inverted~\cite{galbally2013iris}. We achieve similar results with \resist{}-G. Reconstruction accuracy across all three metrics is high. The Type-1 reconstruction accuracy for \resist{}-G is 100\%. This means that every iris in the test set reconstructed using \resist{}-G matched with its real counterpart with a Hamming distance below the threshold. Type-2 and Rank-1 accuracy is comparable to accuracy numbers for the true dataset. 

\begin{table}[t]
\footnotesize
		\centering
		\begin{tabular}{| l |l|r|r|r|}
			\hline
			&& \multicolumn{2}{c|}{TAR} &\\
			Pipeline & Method  &Type1&Type2& Rank-1\\
			\hline
			\multirow{2}{*}{DenseNet} & Legitimate &100\%&98.7\%&99.9\%\\
			& \resist{}-D&75.9\%&43.3\%&{96.3\%}\\
			& Core &30.3\%&18.5\%&{62.2\%}\\

			\hline
        DenseNet & Legitimate &100\%&95.2\%&98.7\%\\
		Normal	& \resist{}-DN&96.7\%&62.5\%&{89.3\%}\\\hline
		\multirow{2}{*}{Gabor} &Legitimate&100\%&88.2\%&97.9\%\\
			& \resist{}-G&100\%&85.2\%&{97.1\%}\\
			\hline
	\end{tabular} 
	\caption{Reconstruction accuracy for 3 different pipelines.  Legitimate shows the accuracy of the underlying pipeline, while the second row shows accuracy for \resist{} images.  True acceptance rate (TAR) is at 1.0\% FAR.}
	\label{tab:resist}
\end{table}
	
\subsection{Discussion} Reconstruction accuracy is proportional to recognition accuracy of a model.  If all the important features are being captured by the feature extractor for comparison, then these features are also available for inversion. 

Type-1 error validates \resist{} networks learning features of a particular image while Type-2 error validates if the learnt features can accurately reconstruct irises of a specific person. The gap in these errors can be attributed to the variance among the irises of a single person.

The \resist{}-D template is the hardest to invert.  We theorize this is due to limited spatial correlation between irises as the feature extraction network takes as input segmented only iris images. Only iris regions are present in the image which are hard to invert.
The \iriscode{} is the easiest to invert since the template contains texture information of the original iris. Occlusions in the \iriscode{} are well defined and are helpful in the inversion process. Among the deep templates normalized iris based templates are easier to invert than non-normalized iris based templates. We attribute this to spatial correlation in the irises and the presence of occlusions. Some deep learning feature extractors use masks per iris image~\cite{zhao2017towards}, we do not use any masks in our work. We hypothesize that masks would provide additional information to the reconstruction. We believe this would improve both recognition and reconstruction accuracy.

Type-1 accuracy for \resist{}-DN is higher than the Type-2 original accuracy curve in Figure~\ref{fig:roc} after $1\%$ percent FAR. We attribute this to sharp reconstructions since the reconstruction network behaves both as a reconstruction and super resolution network due to the presence of perceptual loss.

%
%

%
%
%
%
%
%
%
%

\subsection{Cross Data Set Analysis}
\label{ssec:cross data set analysis}

\begin{table}[t]
\footnotesize
		\centering
		\begin{tabular}{| l |l|r|r|r|r|}
			\hline
			 Dataset  &Type1&Type2& Rank-1&Test accuracy\\
			\hline
			IITD &0\%&0\%&14.6\%&99.9\%\\
			Casia v4 Interval &0\%&0\%&9.45\%&99.5\%\\
            \hline
	\end{tabular} 
	\caption{Reconstruction results when testing \resist{}-D on templates from IITD and Casia v4 Interval images.}
	\label{tab:crossdataset}
\end{table}

To understand the degree to which \resist{} learns general iris structures, we see if \resist{} can  invert irises from other datasets.   For these experiments we focus on the DenseNet model without normalization. We do not perform any retraining on the underlying feature extractor.
 
The two datasets we discuss here are the IITD~\cite{kumar2010comparison} and Casia v4 Interval datasets~\cite{casia}.
Both the IITD and Casia datasets are comprised of 8-bit grayscale iris images (as is the ND-0405 dataset).
The IITD dataset is comprised of 200x200 dimension images (2232 samples) and the Casia dataset is comprised of 300x300 dimension images (2640 samples). Only the right iris images were used for the testing of both the feature extractor and \resist{}-D.  
The feature extractors were not retrained during the cross dataset analysis.
Table~\ref{tab:crossdataset} shows the resulting reconstruction accuracy and performance of the underlying feature extractor.  First note that the DenseNet feature extractor achieves high accuracy without retraining.  However, \resist{}-D has very low accuracy.  

\begin{figure*}[t]
 \begin{subfigure}[b]{0.48\textwidth}
         \centering
          \includegraphics[scale = 0.67]{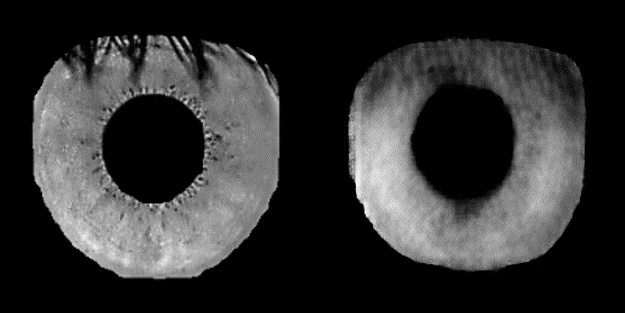}
         \caption{IITD Dataset}
         \label{fig:IITD}
     \end{subfigure}
     \hfill
     \begin{subfigure}[b]{0.40\textwidth}
         \centering
         \includegraphics[scale = 0.55]{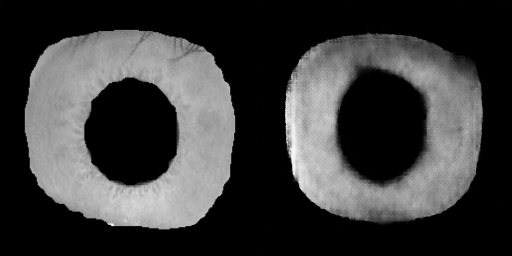}
    \caption{Casia v4 Interval Dataset}
    \label{fig:casia}
     \end{subfigure}
   \caption{Reconstructed irises for other datasets.  Left image is original image, right is reconstructed.}
\end{figure*}

We believe that there are multiple reasons for this performance that result from differences between the ND-0405 dataset and the test datasets.  While the dimensions of the image are similar between datasets the actual size of the iris differs across datasets. Figures~\ref{fig:IITD} and~\ref{fig:casia} show representative images of the reconstructed irises. \resist{}-D recovers the shape of the iris well but does not recover fined grained details. We believe this is represented by a meaningful Rank-1 accuracy but no Type1 or Type2 matches.  Lastly, we believe that \resist{}-D is overfitting to the training samples.  We provide further evidence of this in  Section~\ref{sec:defense}.

\section{Defenses}
\label{sec:defense}
 Regularization is known to help in defending against model inversion attacks~\cite{yang2020defending,shokri2017membership} which is similar to our reconstruction attack. In a model inversion attack, a prediction vector is inverted to a corresponding image where a prediction vector is a $n$-dimensional vector where $n$ is equal to the number of classes in the training/test set. We invert feature vectors~(templates) to images. 
The recognition models (DenseNet and DenseNet Normal) we attack using \resist{} attack models are already regularized using dropout.
Thus, Section~\ref{sec:results} shows how \resist{} performs on a dropout-regularized model. 

We further show reconstruction accuracy when our DenseNet model parameters are regularized using L2 regularization. Essentially this adds another parameter to the loss function of the model to avoid overfitting. The amount of regularization is controlled by $\lambda$.  For this section we restrict attention to DenseNet and \resist{}-D.  L2 regularization does slightly decrease accuracy of the DenseNet model as shown in Test Accuracy column in Table~\ref{tab:crossdataset}. 

Our results for defending against reconstruction attacks are promising, even a small amount of regularization (1e-4) drops the rank-1 reconstruction accuracy to 20\% from 96\%.  This has a minimal impact on the accuracy of the feature extractor. Regularization pushes the feature vectors together such that they have small distances between them. Although the testing accuracy remains unaffected, the \resist{} model struggles with such additional regularization. We believe this  low reconstruction accuracy is a result of an overfitting \resist{} attack model. This was similarly indicated in performing cross-data set analysis (Section~\ref{ssec:cross data set analysis}).

Overfitting of the \resist{} model could be mitigated using various techniques such as dropout and increasing training data in the \resist{} model.  The \resist{} model was delicate and we were not able to defend well against our defense.  We leave this as important future work.

\begin{table}[t]
\footnotesize
		\centering
		\begin{tabular}{| l |l|r|r|r|}
			\hline
			 $\lambda$ L2-regularization  &Type1&Type2& Rank-1& Test Accuracy\\
			\hline
			0 &75.9\%&43.3\%&96.3\% &99.9\% \\
			1e-4 &0\%&0\%&20.4\% &99.7\% \\
			1e-2 &0\%&0\%&0.4\% &99.5\% \\
            \hline
	\end{tabular} 
	\caption{Effect of L2 regularization on effectiveness of $\resist{}$-D.}
	\label{tab:crossdataset}
\end{table}


\section{Conclusion}
\label{sec:conclusion}
We study inverting three types of templates, \iriscode s where prior work exists and has good inversion results, as well as deep templates where there is limited prior work. Our iris reconstruction networks perform well on all three types of templates, with deep templates proving harder to invert than \iriscode. Within the two deep templates segmented only iris templates are harder to invert than normalized iris based templates. 

Our work reinforces the value of template protection mechanisms.  In particular, it seems important to find template protection mechanisms that work on real valued feature vectors where comparison is via the cosine distance. Future work should look into template inversion across datasets, inversion of binary vectors produced from deep neural networks, and the feasibility of storing templates which have limited spatial correlation with their corresponding biometrics. Template inversion in the presence of template protection such as encryption or random sampling is also another direction for future work.

\section*{Acknowledgements}
The authors thank the reviewers for their valuable help in improving the manuscript. This work was supported in part by Synchrony Financial Inc., NSF Grant 1849904, and the McNair scholar program.

\bibliographystyle{IEEEtran}
\bibliography{main}
\balance

\end{document}